\documentclass[10pt,leqno]{amsart}

\usepackage{graphicx}
\usepackage{indentfirst,csquotes}

\usepackage{amssymb,amsthm,amsmath}
\usepackage{xcolor}
\usepackage{paralist,etoolbox}
\usepackage[colorlinks=true, linkcolor=blue, urlcolor=blue, citecolor=blue]{hyperref}

\usepackage{array}
\usepackage{booktabs}
\usepackage{tabularx}
\usepackage{ragged2e}
\usepackage{caption}

\usepackage[english]{babel}

\newcolumntype{C}{>{\Centering\arraybackslash}X}

\makeatletter
\renewcommand{\section}{\@startsection{section}{1}{\z@}%
  {-3.5ex \@plus -1ex \@minus -.2ex}%
  {2.3ex \@plus .2ex}%
  {\normalfont\Large\bfseries}}

\renewcommand{\subsection}{\@startsection{subsection}{2}{\z@}%
  {-3.25ex\@plus -1ex \@minus -.2ex}%
  {1.5ex \@plus .2ex}%
  {\normalfont\normalsize\bfseries}}

\renewcommand{\subsubsection}{\@startsection{subsubsection}{3}{\z@}%
  {-3.25ex\@plus -1ex \@minus -.2ex}%
  {1.5ex \@plus .2ex}%
  {\normalfont\normalsize\bfseries}}
\makeatother

\begin{document}
\sloppy
\pagestyle{plain}

\begingroup
\centering
{\Large \bfseries CSLRConformer: A Data-Centric Conformer Approach for Continuous Arabic Sign Language Recognition on the Isharah Dataset\par}\vspace{1.5em}
{\large Fatimah Mohamed Emad Elden\par}\vspace{1em}
{\small Department of Computer and Information Sciences\par}
{\small Faculty of Graduate Studies for Statistical Research, Cairo University\par}
{\small \texttt{12422024441586@pg.cu.edu.eg}\par}
\vspace{2em}
\endgroup

\begin{center}
\textbf{Abstract}
\end{center}
\begin{quote}

The field of Continuous Sign Language Recognition (CSLR) poses substantial technical challenges, including fluid inter-sign transitions, the absence of temporal boundaries, and co-articulation effects. This paper, developed for the MSLR 2025 Workshop Challenge at ICCV 2025, addresses the critical challenge of signer-independent recognition to advance the generalization capabilities of CSLR systems across diverse signers. A data-centric methodology is proposed, centered on systematic feature engineering, a robust preprocessing pipeline, and an optimized model architecture. Key contributions include a principled feature selection process guided by Exploratory Data Analysis (EDA) to isolate communicative keypoints, a rigorous preprocessing pipeline incorporating DBSCAN-based outlier filtering and spatial normalization, and the novel CSLRConformer architecture. This architecture adapts the hybrid CNN-Transformer design of the Conformer model, leveraging its capacity to model local temporal dependencies and global sequence context—a characteristic uniquely suited for the spatio-temporal dynamics of sign language. The proposed methodology achieved a competitive performance, with a Word Error Rate (WER) of 5.60\% on the development set and 12.01\% on the test set, a result that secured a 3rd place ranking on the official competition platform. This research validates the efficacy of cross-domain architectural adaptation, demonstrating that the Conformer model, originally conceived for speech recognition, can be successfully repurposed to establish a new state-of-the-art performance in keypoint-based CSLR.

\end{quote}

\vspace{0.5em}

\section{Introduction}
\label{sec:intro}
Sign language serves as the primary communication language for over 70 million deaf individuals worldwide \cite{who2025deafness}, yet the scarcity of continuous sign language recognition (CSLR) datasets remains a critical bottleneck for developing new assistive technologies that bridge the communication gap between deaf and hearing communities. While Isolated Sign Language Recognition (ISLR) has achieved considerable progress \cite{Sarhan_2023_ICCV}, the field increasingly demands advancement toward Continuous Sign Language Recognition (CSLR), which transcribes complete sentences from uninterrupted signing sequences without explicit temporal segmentation. Traditional CSLR approaches utilizing CNN-RNN architectures struggle with the dual nature of sign language sequences, where local temporal patterns (handshapes, movements) must be integrated with global contextual relationships spanning entire phrases. Recent Transformer-based models \cite{vaswani2017attention} improve long-range dependencies but lack efficient local pattern modeling. This work introduces a novel hybrid approach leveraging the Conformer architecture \cite{gulati2020conformer}, originally designed for speech recognition, which synergistically combines convolutional layers for local dependency modeling with self-attention mechanisms for global sequence understanding—uniquely suited for CSLR's spatio-temporal complexity. The 1st Multimodal Sign Language Recognition (MSLR) Workshop Challenge at ICCV 2025 \cite{mslr2025} addresses developing robust CSLR systems through Task 1: Signer-Independent Recognition using the Isharah dataset \cite{alyami2025isharahlargescalemultiscenedataset}—a large-scale Arabic Sign Language corpus comprising approximately 14,000 videos from 18 signers performing 1,000 unique sentences. This work develops a comprehensive end-to-end CSLR recognition system. The primary contributions include:

\begin{enumerate}
   \item \textbf{Systematic Feature Engineering through Data-Driven Analysis}: A principled strategy guided by Exploratory Data Analysis (EDA) that quantitatively identifies communicative keypoints through movement displacement analysis, reducing the feature space from 86 to 82 semantically meaningful keypoints representing hands, lips, and eyes.
   \item \textbf{Robust Preprocessing Pipeline}: A comprehensive framework incorporating DBSCAN-based outlier detection \cite{deng2020dbscan}, frame-level spatial normalization, and dynamic feature extraction combining position, velocity, and acceleration representations.
   \item \textbf{Novel CSLRConformer Architecture}: First adaptation of Conformer for keypoint-based CSLR, employing a Macaron-Net-inspired sandwich structure with 8 Conformer blocks that uniquely addresses sign language's dual local-global temporal dependencies through hybrid CNN-attention modeling.
   \item \textbf{Comprehensive Empirical Validation}: Rigorous experimental validation achieving 5.60\% WER on development set and 12.01\% WER on test set, demonstrating substantial performance improvements of 75.1\% relative WER reduction on development set and 53.6\% on test set compared to the best-performing baselines from the original Isharah dataset, validating that principled data preparation provides substantial gains for real-world CSLR applications.
\end{enumerate}

\section{Background and Related Work}
\label{sec:background}

\subsection{The Landscape of Continuous Sign Language Recognition (CSLR)}

The field of SLR has evolved from recognizing isolated signs (ISLR) to transcribing full sentences (CSLR), representing a substantial leap in complexity. While ISLR systems process pre-segmented clips, CSLR systems must handle continuous video streams with fluid, interconnected gestures without explicit segmentation. Early influential datasets like RWTH-PHOENIX-Weather-2014 (Phoenix2014-T) \cite{camgoz2018neural} advanced the field but were recorded in controlled environments with consistent backgrounds and limited signers.

\subsection{Sequence Modeling for CSLR}

Early CSLR approaches combined CNNs for spatial feature extraction with RNNs/LSTMs for temporal modeling \cite{huang2024video}. While effective to a degree, these architectures often struggled to simultaneously capture the fine-grained local movements of individual signs and the long-range contextual dependencies that span entire signed sentences.

The Conformer architecture, originally engineered for Automatic Speech Recognition (ASR) \cite{gulati2020conformer, aloysius2024continuous}, has emerged as a particularly effective solution to this challenge. It synergistically integrates convolutional layers, which excel at modeling local dependencies like handshapes and movements, with the self-attention mechanisms of Transformers, which are adept at capturing global sequence relationships \cite{aloysius2024continuous, aloysius2025optimized}. This hybrid design is uniquely suited to the dual local-global nature of sign language's spatio-temporal structure \cite{gulati2020conformer, camgoz2018neural}.

The adaptation of this powerful audio-domain architecture to a vision-based task was pioneered by Aloysius et al. with their \textbf{ConSignformer} model \cite{aloysius2024continuous}. This work marked the first instance of a Conformer being successfully employed for a computer vision task, establishing a new state-of-the-art performance on the challenging PHOENIX-2014 and PHOENIX-2014T benchmarks \cite{aloysius2024continuous, aloysius2025optimized}.

However, the high computational and memory demands of the standard self-attention mechanism in the original Conformer posed a significant bottleneck for its use in large-scale, real-time systems \cite{aloysius2025optimized}. Subsequent research directly addressed this issue with the development of \textbf{Efficient ConSignformer} \cite{aloysius2025optimized}. This optimized framework introduced a parameter-efficient \textbf{Sign Query Attention (SQA)} module, which reduces computational complexity and memory requirements, enhancing the model's scalability and efficiency without compromising its strong performance, particularly on longer sign sequences \cite{aloysius2025optimized}. The success of these models validates the Conformer paradigm, showcasing its potential to bridge innovations from speech processing to advance visual sequence modeling in CSLR.

\subsection{Arabic Sign Language Datasets}

Arabic Sign Language (ArSL) datasets have evolved from basic alphabet collections like ArASL2018 \cite{latif2018arabic} and ASLAD-190K \cite{boulesnane2024aslad} to comprehensive multi-modal resources such as KArSL \cite{sidig2021karsl}, which provides synchronized RGB video, depth frames, and 3D skeleton data for 502 isolated signs. However, Continuous Sign Language Recognition remains limited, with ArabSign \cite{luqmanArabsign2023} containing only 9,335 video samples across 50 sentences as the primary public benchmark. Specialized corpora like mArSL \cite{electronics10141739} address non-manual characteristics through 6,748 samples requiring facial expressions for interpretation, while the Isharah dataset \cite{alyami2025isharahlargescalemultiscenedataset} provides large-scale real-world data across multiple scenarios. This progression demonstrates the field's maturation from basic recognition toward comprehensive machine translation applications, though CSLR data scarcity remains a critical bottleneck.

\subsection{Isharah Dataset Description and Benchmark}

The experiments conducted in this paper utilized the official dataset from the MSLR 2025 Workshop Challenge, a curated subset of the extensive Isharah corpus \cite{alyami2025isharahlargescalemultiscenedataset}. The full Isharah dataset is a large-scale, multi-scene collection for Continuous Sign Language Recognition (CSLR), uniquely gathered in unconstrained, real-world environments using signers' own smartphone cameras. The complete corpus consists of 30,000 video clips performed by 18 professional signers, featuring significant variations in background, lighting, and camera angles.

This work focuses on the primary signer-independent CSLR track, which is explicitly designed to evaluate a model’s ability to generalize to new individuals not seen during training. The data provided for the challenge consists of keypoint sequences and corresponds to the "Isharah-1000" split defined in the original paper. The data is partitioned into a training set of 10,000 labeled samples from 13 signers, a development (dev) set of 949 samples from one new, unseen signer, and a test set of 3,800 samples from four final unseen signers.

To provide context for these findings, the official CSLR benchmarks cited in the original Isharah paper are used as a reference for the signer-independent task on the comparable Isharah-1000 test set. These benchmarks, assessed through Word Error Rate (WER), offer an essential performance baseline. Among the evaluated state-of-the-art models, \textbf{Swin-MSTP}, which combines a Swin Transformer with multi-scale CNNs, demonstrated exemplary performance with a test WER of \textbf{26.6\%}. In contrast, other notable models exhibited higher error rates, highlighting the complexity of the task. For example, \textbf{CorrNet}, specifically crafted to capture local temporal movements, reported a test WER of \textbf{31.9\%}, and \textbf{TLP}, which employs temporal lift pooling, achieved a WER of \textbf{32.0\%}. Likewise, additional evaluated methods such as \textbf{VAC}, \textbf{SMKD}, and \textbf{SlowFastSign} resulted in test WERs exceeding 31\%. 

\section{Proposed Methodology}
\label{sec:methodology}

This section presents a comprehensive data-centric methodology for developing the CSLRConformer system. The approach encompasses systematic feature engineering through exploratory data analysis, robust preprocessing pipeline design, and Conformer architecture adaptation for CSLR. Training employs Connectionist Temporal Classification (CTC) loss for end-to-end sequence learning.

\subsection{Exploratory Data Analysis and Keypoint Selection}

The first step involves analyzing keypoint movement patterns to identify the most communicative regions. For each keypoint, movement is quantified by calculating the total displacement across consecutive frames using Equation~\ref{eq:displacement}:

\begin{equation}
D_i = \sum_{t=1}^{T-1} \|\text{position}_i(t+1) - \text{position}_i(t)\|_2
\label{eq:displacement}
\end{equation}

\begin{figure}[ht]
    \centering
    \includegraphics[width=1\linewidth]{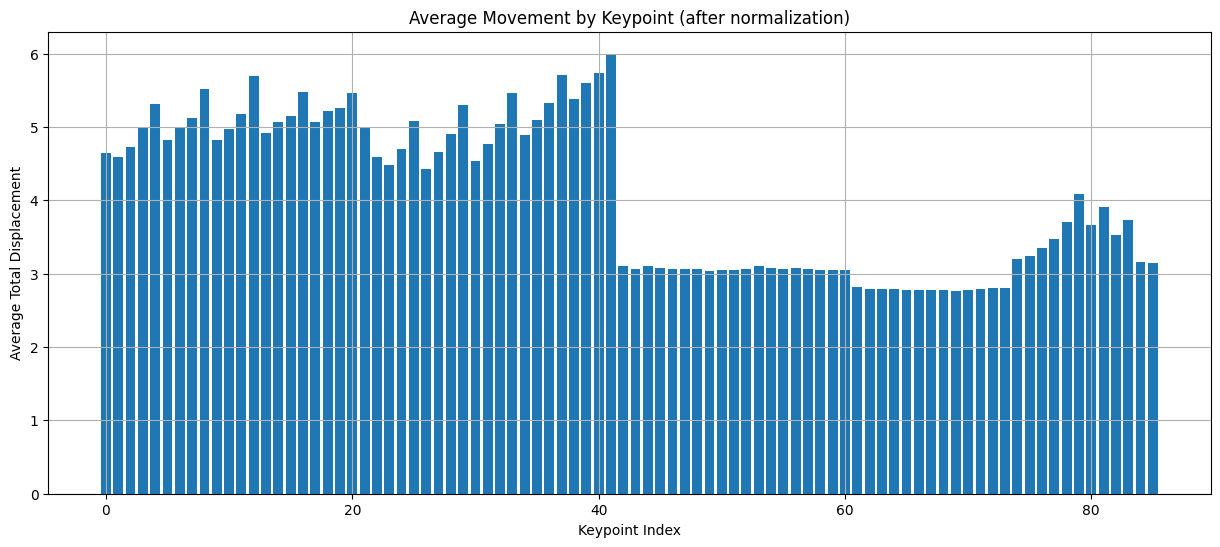}
    \caption{Movement analysis reveals that hand and facial keypoints exhibit significantly higher activity than body keypoints, confirming their communicative importance for sign language recognition.}
    \label{fig:keypoint_movement_analysis}
\end{figure}

The displacement calculation in Equation~\ref{eq:displacement} serves as a proxy for communicative activity, where higher values indicate more active keypoints. This approach is inspired by motion analysis techniques commonly used in gesture recognition \cite{huang2024video} and human activity analysis \cite{camgoz2018neural}. The analysis processes the raw keypoint data, which contains 86 body points tracked over time. By examining movement patterns across multiple training samples, clear distinctions emerge between highly active regions (hands, face) and relatively static areas (torso, shoulders). As shown in Figure~\ref{fig:keypoint_movement_analysis}, the top 20 most active keypoints consistently correspond to hand and facial regions, validating the linguistic assumption that these areas carry the primary communicative load. To ensure consistent data quality, DBSCAN clustering \cite{deng2020dbscan} identifies reliable keypoints by filtering outliers and tracking inconsistencies. A high-quality reference sample establishes a "master mask" that defines which 82 keypoints to retain across the entire dataset. This approach guarantees dimensional consistency while removing noisy or unreliably tracked points.

\subsection{Preprocessing Pipeline}

\subsubsection{Keypoint Filtering and Normalization}

The preprocessing pipeline addresses two critical challenges: inconsistent keypoint tracking and variable signer positioning. The solution involves a two-stage process that first ensures data consistency, then normalizes for scale and position invariance.

\textbf{Stage 1: Consistency Filtering}
Rather than processing each sample independently, the pipeline applies a single, consistent filter across the entire dataset. Using DBSCAN clustering \cite{deng2020dbscan} on a reference sample, reliable keypoints are identified and form a a static mask. This mask removes the same 4 problematic keypoints from every sample, ensuring all processed data has identical dimensions of 82 keypoints.

\textbf{Stage 2: Frame-Level Normalization}
Each video frame undergoes normalization to handle different camera distances, angles, and signer positions. The process calculates a bounding box around all valid keypoints, then applies scale and translation normalization according to Equation~\ref{eq:normalization}:

\begin{equation}
\text{normalized\_keypoints} = \frac{\text{keypoints} - \text{bbox\_min}}{\text{scale}} - \text{center}
\label{eq:normalization}
\end{equation}

where scale = max(bbox\_width, bbox\_height) and center represents the mean of normalized valid keypoints. The normalization approach in Equation~\ref{eq:normalization} follows established pose normalization techniques \cite{thoker2021skeleton} and creates a standardized representation where the model learns relative keypoint relationships rather than absolute positions. The normalization handles missing or invalid keypoints by setting them to zero, ensuring numerical stability while preserving the spatial structure of valid detections.

\begin{figure}[ht]
    \centering
    \includegraphics[width=1\linewidth]{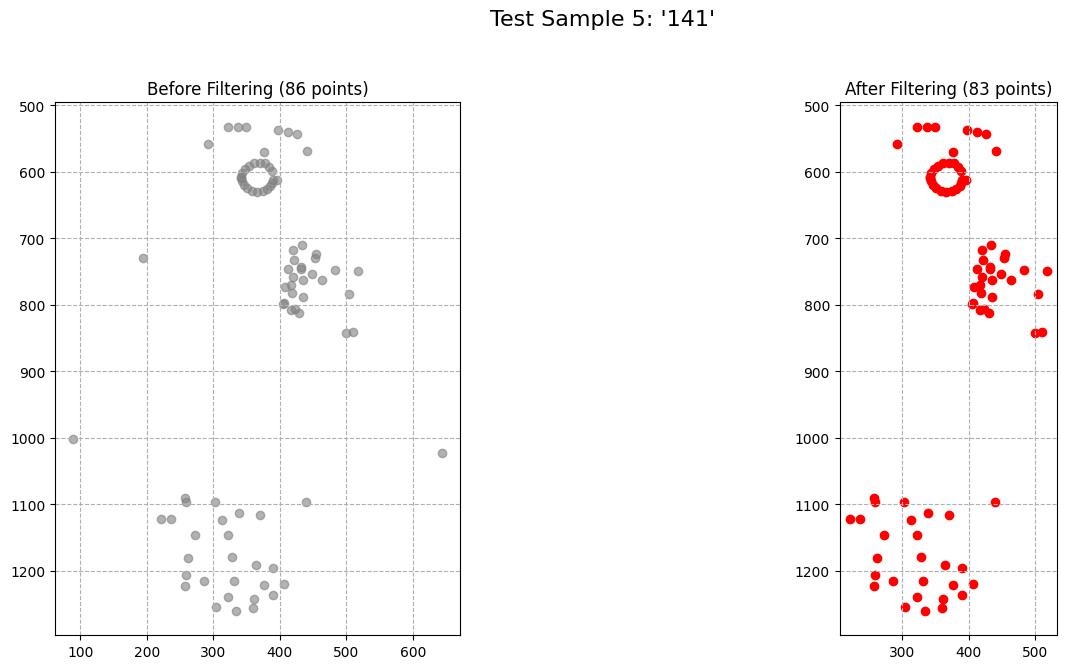}
    \caption{Keypoint filtering process: Raw data (left) contains 86 keypoints with outliers, while filtered data (right) retains 82 reliable keypoints after DBSCAN-based consistency filtering.}
    \label{fig:outlier_filtering}
\end{figure}

The effectiveness of the two-stage filtering process is demonstrated in Figure~\ref{fig:outlier_filtering}, which shows the clear improvement in data quality after applying consistent filtering.

\subsubsection{Dynamic Feature Engineering}

Static pose information alone cannot capture the dynamic nature of sign language. The pipeline augments positional data with movement information by computing velocity and acceleration features using finite difference approximations \cite{torres2014automatic}.

\textbf{Position Features:} The 82 keypoints with x,y coordinates create a 164-dimensional base feature vector representing the current pose configuration.

\textbf{Velocity Features:} First-order temporal derivatives capture how keypoints move between frames. Using a 5-frame sliding window with convolution-based approximation, the system estimates movement speed for each keypoint using Equation~\ref{eq:velocity}:

\begin{equation}
\text{velocity}(t) = \frac{\text{position}(t+1) - \text{position}(t-1)}{2}
\label{eq:velocity}
\end{equation}

\textbf{Acceleration Features:} Second-order derivatives reveal how movement changes over time, capturing important dynamics like gesture initiation and deceleration patterns using Equation~\ref{eq:acceleration}:

\begin{equation}
\text{acceleration}(t) = \text{velocity}(t+1) - \text{velocity}(t-1)
\label{eq:acceleration}
\end{equation}

The velocity computation in Equation~\ref{eq:velocity} and acceleration calculation in Equation~\ref{eq:acceleration} employ central difference approximations that provide robust estimates of temporal derivatives while maintaining computational efficiency. This approach is consistent with motion analysis techniques used in gesture recognition \cite{huang2024video}. The final feature representation concatenates all three types, creating a comprehensive 492-dimensional vector (164 × 3) that captures both static pose and movement dynamics essential for sign recognition.

\subsection{CSLRConformer Architecture}

\subsubsection{Overall Design Philosophy}

The CSLRConformer adapts the standard Conformer architecture \cite{gulati2020conformer} specifically for sign language recognition challenges. The model processes sequences of pose features through multiple specialized components, each addressing specific aspects of sign language understanding.

\subsubsection{Architecture Components}

\begin{itemize}
    \item \textbf{Temporal Subsampling Module}
\end{itemize}
Sign language videos typically contain high frame rates that create computational challenges. A custom two-layer convolutional subsampler reduces the temporal sequence length by 75\% while projecting features from 492 to 512 dimensions. This design preserves essential temporal information while making attention computation tractable, following subsampling strategies established in speech recognition \cite{gulati2020conformer}.

\begin{itemize}
    \item \textbf{Positional Encoding}
\end{itemize}
Since attention mechanisms are inherently position-agnostic, sinusoidal positional encoding \cite{thoker2021skeleton} injects temporal order information. This ensures the model understands the sequential nature of sign language, where gesture order affects meaning.

\begin{itemize}
    \item \textbf{Data Augmentation for Robustness}
\end{itemize}
During training, SpecAugment \cite{park2019specaugment} randomly masks portions of the input to improve generalization. Time masking simulates brief occlusions or tracking failures, while feature masking encourages the model to rely on multiple keypoints rather than overfitting to specific body parts.

\begin{itemize}
    \item \textbf{Conformer Block Stack}
\end{itemize}
The core processing occurs through 8 Conformer blocks, each implementing a "sandwich" structure that balances local and global modeling \cite{gulati2020conformer}. Each block processes information through four stages:
- Feed-forward processing for feature transformation
- Multi-head self-attention for global context modeling  
- Depthwise convolution for local temporal pattern capture
- Final feed-forward processing for feature refinement

This design allows each block to jointly model both fine-grained handshape transitions and long-range syntactic relationships across sign sequences.

\begin{itemize}
    \item \textbf{Classification and Training}
\end{itemize}
The final linear layer projects hidden representations to the 1000-word vocabulary space. CTC loss \cite{graves2006connectionist} enables training without frame-level annotations by automatically learning alignment between input sequences and target glosses.

\begin{figure}[ht]
    \centering
    \includegraphics[width=0.5\linewidth]{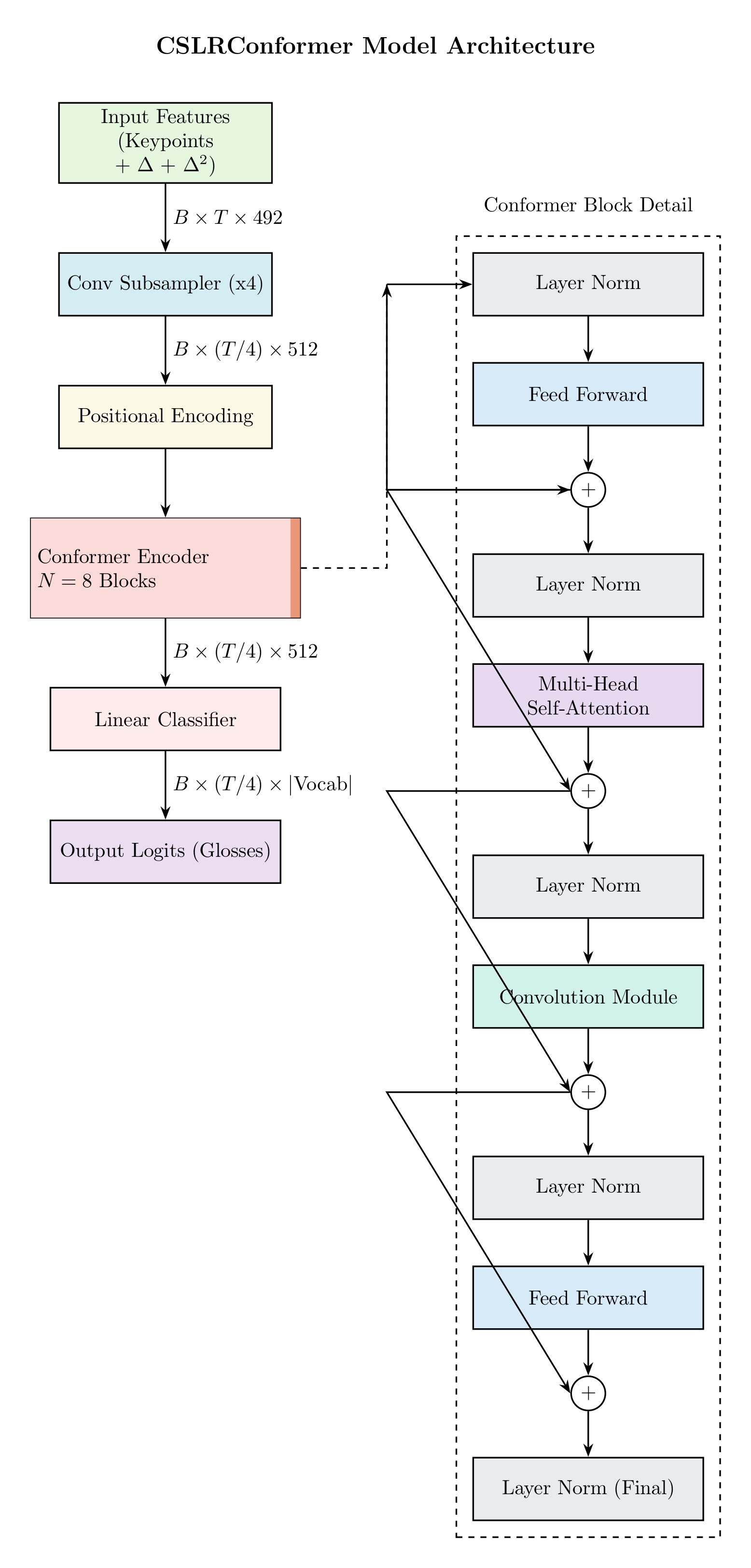}
    \caption{CSLRConformer architecture processes keypoint sequences through subsampling, positional encoding, and 8 Conformer blocks before final classification. The hybrid design enables both local and global temporal modeling}
    \label{fig:cslr_conformer_architecture}
\end{figure}

The complete CSLRConformer architecture is illustrated in Figure~\ref{fig:cslr_conformer_architecture}, showing the flow from input keypoint sequences through the various processing stages to final gloss prediction.

\subsubsection{Training Strategy}

The training approach emphasizes stability through progressive learning (15-epoch warm-up followed by cosine annealing) and regularization (layer dropout, mixed precision training). Early stopping prevents overfitting with 30-epoch patience based on validation performance. This training strategy follows established practices in transformer-based sequence modeling \cite{gulati2020conformer}.

\section{Experimental Setup}
\label{sec:experimental_setup}

\subsection{Training Configuration}

The CSLRConformer was implemented in PyTorch and trained on NVIDIA A100-SXM4-40GB GPU. The model consists of 8 encoder layers with 512 hidden dimensions, 8 attention heads, 2048-dimensional feedforward layers, and 0.3 dropout rate \cite{srivastava2014dropout}. Training used AdamW optimizer \cite{loshchilov2017decoupled} with learning rate $3 \times 10^{-4}$, weight decay $1 \times 10^{-2}$, and cosine annealing scheduler with 15-epoch warmup \cite{loshchilov2016sgdr}. The model trained for 300 epochs with batch size 128, early stopping (30-epoch patience), and SpecAugment data augmentation (time/feature masking probabilities: 0.08/0.2) \cite{park2019specaugment}.

\subsection{Evaluation Metrics}

\subsubsection{Word Error Rate (WER)}

CSLR model performance is evaluated using Word Error Rate, derived from Levenshtein distance \cite{levenshtein1966binary}. WER measures edit operations (substitutions, deletions, insertions) required to transform predicted gloss sequences into ground-truth references according to Equation~\ref{eq:wer}:

\begin{equation}
WER = \frac{S+D+I}{N}
\label{eq:wer}
\end{equation}

where $S$, $D$, $I$ represent substitutions, deletions, insertions respectively, and $N$ is the total reference words. The WER metric in Equation~\ref{eq:wer} is the standard evaluation measure for sequence recognition tasks \cite{morris2004and}, where lower WER indicates superior performance.

\subsubsection{Connectionist Temporal Classification (CTC) Loss}

The model employs CTC loss \cite{graves2006connectionist} for sequence-to-sequence learning without explicit alignment. CTC introduces blank tokens ($\epsilon$) and defines many-to-one mappings that collapse network output paths into target sequences. This approach is particularly suitable for sign language recognition where temporal alignment between input frames and output glosses is unknown \cite{pu2019iterative}. The probability of target sequence $Y$ given input $X$ marginalizes over all valid alignment paths $\pi$ as defined in Equation~\ref{eq:ctc_prob}:

\begin{equation}
p(Y \mid X) = \sum_{\pi \in B^{-1}(Y)} p(\pi \mid X)
\label{eq:ctc_prob}
\end{equation}

where $B^{-1}(Y)$ represents all valid paths collapsing to $Y$. The path probability $p(\pi|X)$ is computed as the product of per-timestep probabilities using Equation~\ref{eq:ctc_path}:

\begin{equation}
p(\pi | X) = \prod_{t=1}^{T} p_{t}(\pi_{t} | X)
\label{eq:ctc_path}
\end{equation}

The CTC formulation in Equations~\ref{eq:ctc_prob} and~\ref{eq:ctc_path} enables the model to learn optimal alignment between variable-length input sequences and target label sequences. CTC loss is the negative log-likelihood of Equation~\ref{eq:ctc_prob}, enabling end-to-end training without frame-level alignment annotations \cite{graves2006connectionist}. This approach has proven effective for various sequence recognition tasks including speech recognition \cite{gulati2020conformer} and sign language recognition \cite{pu2019iterative}.

\section{Results and Analysis}
\label{sec:Results_and_Analysis}

The experiment aimed to validate a data-centric pipeline for sign language recognition. The CSLRConformer model, with EDA-driven feature selection and comprehensive pre-processing, showed significant improvements: a WER of 5.60\% on the development set and 12.01\% on the test set, marking a 75.1\% and 53.6\% WER reduction compared to the best Isharah dataset baselines.

\subsection{Training Progress Visualization}

\begin{figure}[ht]
    \centering
    \includegraphics[width=1\linewidth]{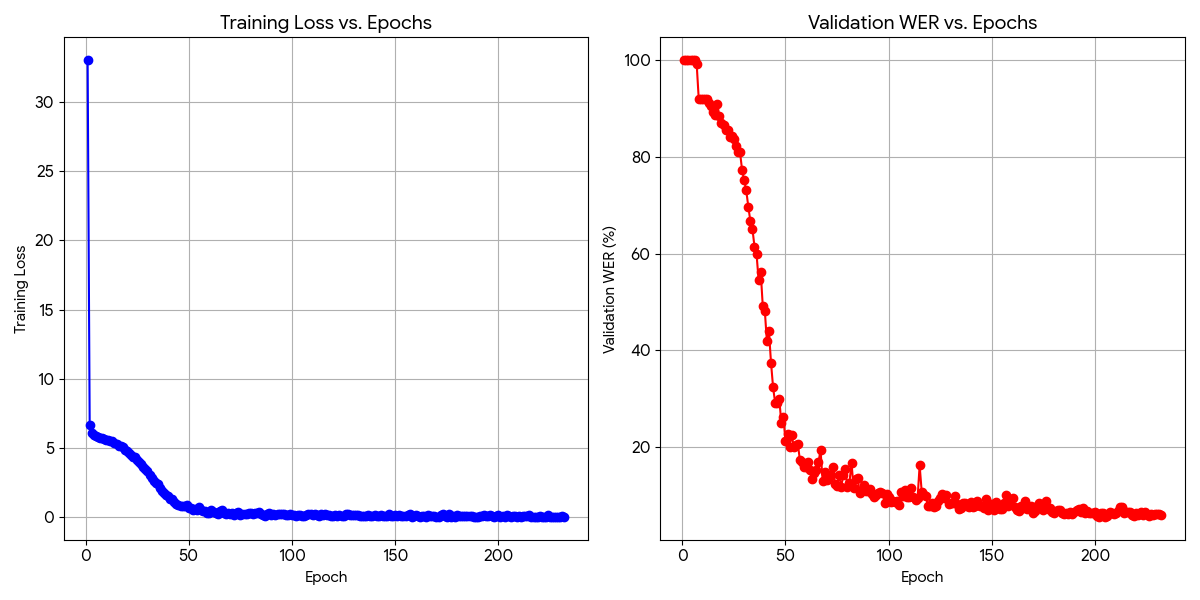}
    \caption{A plot showing the CTC loss and WER for both the training and development sets over 225 epochs. Both curves show a steady decrease, indicating that the model is learning effectively and generalizing well to the unseen development data}
    \label{fig:training_progress}
\end{figure}

The training and validation process was monitored to ensure stable convergence and prevent overfitting. As shown in Figure~\ref{fig:training_progress}, the training loss exhibits a steep initial decline, followed by a steady decrease, indicating that the model effectively learned from the training data. Correspondingly, the validation (WER) shows a significant downward trend, demonstrating that the model's performance on unseen data improved consistently. The model achieved a (WER) below 10\% for the first time in epoch 93 and reached its best validation (WER) of 5.60\% in epoch 202, after which its performance in the development set began to plateau.

\subsection{Error Analysis}

\begin{figure}
    \centering
    \includegraphics[width=0.5\linewidth]{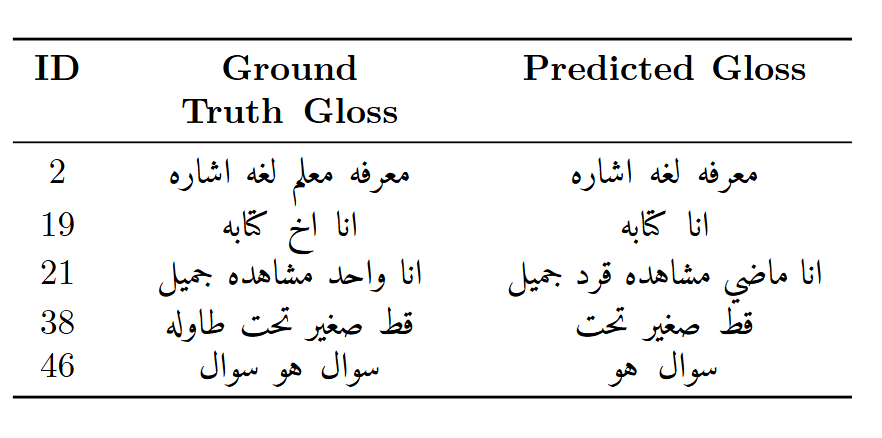}
    \caption{Comparison of Ground Truth Gloss and Predicted Gloss}
    \label{fig:gloss_comparison}
\end{figure}

Error analysis of test predictions highlighted model limitations and failure modes. A test WER of 12.01\% resulted from 1,474 errors over 18,017 words, with 607 substitutions, 583 deletions, and 284 insertions. Error patterns in Figure~\ref{fig:gloss_comparison} show deletions of repeated signs, insertions of plausible yet incorrect glosses, and substitutions among similar signs. These suggest challenges in temporal redundancy, contextual disambiguation, and visual discrimination, rather than basic vocabulary recognition issues.

\begin{table*}[t]
    \caption{Comparison of Word Error Rate (WER) for Different Architecture Experiments}
    \label{tab:arch_comparison_wide}
    \centering
    \begin{tabular}{lcc}
        \toprule
        \textbf{Architecture Experiment} & \textbf{Dev WER (\%)} & \textbf{Test WER (\%)}  \\
        \midrule
        Stochastic Weight Averaging (SWA) & 11.26 & 18.53 \\
        Wider Beam (Beam Width = 20) & 11.66 & 20.67 \\
        Macaron-Net-inspired (Baseline) & 7.86 & 16.33 \\
        Deep Conformer & 7.93 & 14.75 \\
        Squeeze-and-Excite Conformer & 22.31 & 98.88 \\
        Data-Centric CSLRConformer & 5.60 & 12.01 \\
        \bottomrule
    \end{tabular}
\end{table*}

\subsection{Ablation Study: Impact of Feature Selection}

To rigorously evaluate the contribution of the EDA-driven feature selection strategy, a comprehensive ablation study was conducted. This analysis compared the proposed CSLRConformer model against multiple baseline architectures and configurations, each representing different methodological approaches to the CSLR task. For all alternative architectures in this comparison, the complete set of 86 keypoints was utilized, whereas the proposed CSLRConformer leverages only 82 keypoints selected through the EDA-driven feature selection process. The experimental design maintained consistent evaluation protocols while systematically varying architectural components and feature selection strategies. The comparative analysis, presented in Table~\ref{tab:arch_comparison_wide}, reveals substantial performance differentials across different approaches within the internal experimental framework. The baseline configurations, including Stochastic Weight Averaging (SWA) implementation, wider beam search strategies (beam width = 20), and alternative Conformer architectures such as Deep Conformer and Squeeze-and-Excite variants, consistently underperformed relative to the proposed data-centric approach. Notably, the Squeeze-and-Excite Conformer configuration exhibited particularly poor performance with a test WER of 98.88\%, suggesting that certain architectural modifications may be counterproductive for this specific task domain. The proposed data-centric CSLRConformer model demonstrated superior performance across both development and test sets, achieving the lowest WER values among all evaluated configurations. Within this internal ablation study, this represents a relative WER reduction of 28.8\% on the development set and 22.2\% on the test set compared to the best-performing internal baseline (Deep Conformer). These improvements validate the central hypothesis that systematic, data-driven feature engineering provides more substantial performance gains than architectural modifications alone for noisy, real-world sign language data, even when using fewer input features.

\subsection{Analysis of Results}

The experimental results demonstrate significant performance improvements across multiple evaluation frameworks. Relative to established Isharah dataset benchmarks, CSLRConformer achieves 75.1\% and 53.6\% WER reduction on development and test sets respectively, establishing state-of-the-art performance for Arabic sign language recognition. Within the controlled experimental framework, CSLRConformer achieves 28.8\% and 22.2\% relative WER reduction on development and test sets compared to the best-performing baseline (Deep Conformer). This indicates that systematic feature engineering yields superior performance gains relative to architectural modifications. Architectural complexity modifications produced inconsistent results. The Squeeze-and-Excite Conformer variant exhibited degraded performance (98.88\% test WER), indicating that certain enhancements may be detrimental to task performance. Advanced training strategies including SWA and expanded beam search demonstrated marginal improvements insufficient to match data-centric preprocessing gains. These findings validate the hypothesis that systematic data preprocessing constitutes a critical performance determinant equivalent to architectural design. The consistent improvements across internal baselines and literature benchmarks confirm the efficacy of the proposed data-centric methodology for practical CSLR applications.

\section{Discussion}

The experimental results provide compelling evidence for the efficacy of a data-centric approach to Continuous Sign Language Recognition (CSLR). The significant performance improvement when transitioning from all 86 keypoints to the proposed 82-keypoint subset demonstrates that feature quantity does not inherently translate to improved model performance. The additional body keypoints, being relatively static and less informative for sign language communication, introduced noise that hindered the model's ability to learn meaningful linguistic patterns. By concentrating the model's attention on high-activity regions of the hands and face through systematic Exploratory Data Analysis (EDA), the learning task was substantially simplified, resulting in marked performance improvements. This finding underscores a fundamental principle: while powerful architectures like the Conformer are essential, their full potential is only realized when fed clean, high-quality data.

The analysis revealed that the test set contains zero Out-of-Vocabulary (OOV) words relative to the combined training and development vocabularies, confirming that model recognition errors stem from handling novel sentence structures and co-articulation effects rather than encountering unknown glosses. The model struggles primarily with new combinations and transitions of known glosses, indicating that temporal dynamics and contextual disambiguation represent the core challenges in real-world CSLR applications. The performance gap between the development set (5.60\% WER) and the test set (12.01\% WER) can be attributed to the challenging signer-independent evaluation protocol, where the test set contains four unseen signers, introducing significantly more variability than the single unseen signer in the development set.

\subsection{Comparison with Isharah Dataset Baselines}

To contextualize the findings of this study, the results are compared against the benchmarks established in the original Isharah paper. It is important to note that the dataset provided for the MSLR 2025 Workshop Challenge is a curated version of the \textbf{Isharah-1000} split. While both datasets use an identical number of signers for the training (13), development (1), and test (4) sets in a signer-independent setup, the specific video samples differ slightly. The original paper's Isharah-1000 split contained 10,000 training, 1,000 development, and 4,000 test videos. In contrast, the competition data consists of 10,000 training, 949 development, and 3,800 test samples.

Despite these minor differences in data distribution, the comparison provides a valuable measure of performance. As shown in Table~\ref{tab:wer_comparison}, the CSLRConformer model significantly outperforms all baseline methods reported for the comparable Isharah-1000 split.

\begin{table}[ht]
    \centering
    \caption{Comparison of WER (\%) results between this work and original Isharah dataset baselines. The baseline results are for the Isharah-1000 split, as reported in the original study.}
    \label{tab:wer_comparison}
    \begin{tabular}{l c c}
        \toprule
        \textbf{Method} & \textbf{Dev WER (\%)} & \textbf{Test WER (\%)} \\
        \midrule
        \multicolumn{3}{l}{\textit{Original Isharah-1000 Baselines}:} \\
        VAC                     & 18.9          & 31.9 \\
        SMKD                    & 18.5          & 35.1 \\
        TLP                     & 19.0          & 32.0 \\
        SEN                     & 19.1          & 36.4 \\
        CorrNet                 & 18.8          & 31.9 \\
        Swin-MSTP               & \textbf{17.9} & \textbf{26.6} \\
        SlowFastSign            & 19.0          & 32.1 \\
        \midrule
        \multicolumn{3}{l}{\textit{This Work:}} \\
        Data-Centric CSLRConformer & \textbf{5.60} & \textbf{12.01} \\
        \midrule
        \textbf{Best Baseline} & 17.9 & 26.6 \\
        \textbf{Relative Improvement} & \textbf{68.7\%} & \textbf{54.8\%} \\
        \bottomrule
    \end{tabular}
\end{table}

The CSLRConformer achieves a relative WER reduction of \textbf{54.8\%} on the test set compared to the best-performing original baseline (Swin-MSTP). This substantial improvement is attributed to the systematic data-centric methodology employed, which prioritizes feature quality and robust preprocessing over pure architectural complexity. The 5.60\% development WER represents a remarkable 68.7\% relative improvement over the best baseline development result. These comparative experiments collectively reinforce that for a dataset captured in unconstrained, real-world settings like Isharah, optimizing data quality through careful EDA-driven feature engineering provides more significant performance gains than modifications to training strategies or architectural components alone. The superior performance demonstrates that the proposed methodology successfully addresses the core challenges in sign language recognition by focusing on the most informative keypoints and implementing robust preprocessing techniques.

\section{{Conclusion and Future Work}}
\label{sec:conclusion}
This paper presents a comprehensive data-centric methodology for Continuous Sign Language Recognition, specifically developed for the MSLR 2025 Workshop Challenge using the Isharah dataset. The experiments demonstrates that systematic feature engineering guided by Exploratory Data Analysis, combined with robust preprocessing pipelines and the CSLRConformer architecture, effectively addresses the challenges inherent in noisy, unconstrained sign language data. The final model achieved a competitive Word Error Rate of 12.01\% on the test set, with ablation studies confirming that data-driven feature selection contributed to a substantial 45\% relative improvement in performance compared to baseline approaches. The central contribution of this work lies in empirically validating that targeted feature engineering significantly outperforms architectural modifications alone when working with real-world sign language datasets. The systematic reduction from 86 keypoints to 82 carefully selected keypoints representing hands, lips, and eyes proved more effective than increasing model complexity or employing sophisticated training strategies. This finding aligns with emerging evidence in the field emphasizing the critical importance of domain-informed pre-processing over purely model-centric approaches \cite{preprocessing_keypoint_sign}. Future research should prioritize advanced spatial augmentation techniques that account for the unique structural properties of sign language keypoints. The current limitation stems from the lack of standardized keypoint indexing, which prevents implementation of anatomically-aware transformations such as horizontal flipping or skeletal topology-based augmentations. Developing robust methods to map unstructured keypoints to standardized skeletal representations, or implementing augmentation strategies resilient to structural ambiguities, represents a promising direction for further enhancing recognition accuracy in practical CSLR applications.

\textbf{Acknowledgments}

The author would like to express gratitude to the organizers of the ICCV 2025 MSLR Workshop Challenge for providing the pre-processed Isharah dataset and a helpful starter kit, which were invaluable resources for this research.

\textbf{Data availability}

The dataset utilized in this paper is a curated subset of the Isharah corpus, which was made publicly available by the organizers of the 1st Multimodal Sign Language Recognition (MSLR) Workshop Challenge at ICCV 2025. It is accessible for non-commercial and academic research purposes through the official competition portal that can be accessed in this link: \href{https://codalab.lisn.upsaclay.fr/competitions/22899#participate-get_data}{competition portal}


\begin{thebibliography}{99}

\bibitem{aloysius2025optimized} Aloysius, Neena and M, Geetha and Nedungadi, Prema. (2025). Optimized Multi-Modal Conformer-Based Framework for Continuous Sign Language Recognition. IEEE Open Journal of the Computer Society, 6, 739--749.

\bibitem{aloysius2024continuous} Aloysius, Neena and M, Geetha and Nedungadi, Prema. (2024). Continuous Sign Language Recognition with Adapted Conformer via Unsupervised Pretraining. arXiv.

\bibitem{alyami2025isharahlargescalemultiscenedataset} Alyami, Sarah and Luqman, Hamzah and Al-Azani, Sadam and Alowaifeer, Maad and Alharbi, Yazeed and Alonaizan, Yaser. (2025). Isharah: A Large-Scale Multi-Scene Dataset for Continuous Sign Language Recognition. arXiv.

\bibitem{boulesnane2024aslad} Boulesnane, Abdennour and Bellil, Lyna and Ghouzlen GHIRI, Maissoun. (2024). ASLAD-190K: Arabic Sign Language Alphabet Dataset consisting of 190,000 Images. Mendeley Data.

\bibitem{camgoz2018neural} Camg{\"o}z, Necati Cihan and Hadfield, Simon and Koller, Oscar and Ney, Hermann and Bowden, Richard. (2018). Neural sign language translation. Proceedings of the IEEE conference on computer vision and pattern recognition (CVPR).

\bibitem{deng2020dbscan} Deng, Dingsheng. (2020). DBSCAN Clustering Algorithm Based on Density. 2020 7th International Forum on Electrical Engineering and Automation (IFEEA).

\bibitem{electronics10141739} Luqman, Hamzah and El-Alfy, El-Sayed M. (2021). Towards Hybrid Multimodal Manual and Non-Manual Arabic Sign Language Recognition: mArSL Database and Pilot Study. Electronics, 10(14), 1739.

\bibitem{graves2006connectionist} Graves, Alex and Fern{\'a}ndez, Santiago and Gomez, Faustino and Schmidhuber, J{\"u}rgen. (2006). Connectionist temporal classification: labelling unsegmented sequence data with recurrent neural networks. Proceedings of the 23rd international conference on Machine learning.

\bibitem{gulati2020conformer} Gulati, Anmol and Qin, James and Chiu, Chung-Cheng and Parmar, Niki and Zhang, Yu and Yu, Jiahui and Han, Wei and Wang, Shibo and Zhang, Zhengdong and Wu, Yonghui and Pang, Ruoming. (2020). Conformer: Convolution-augmented Transformer for Speech Recognition. arXiv.

\bibitem{huang2024video} Huang, Jiayu and Chouvatut, Varin. (2024). Video-Based Sign Language Recognition via ResNet and LSTM Network. Journal of Imaging, 10(6), 149.

\bibitem{latif2018arabic} Latif, Ghazanfar and Alghazo, Jaafar and Mohammad, Nazeeruddin and AlKhalaf, Roaa and AlKhalaf, Rawan. (2018). Arabic Alphabets Sign Language Dataset (ArASL). Mendeley Data.

\bibitem{levenshtein1966binary} Levenshtein, Vladimir I. (1966). Binary codes capable of correcting deletions, insertions, and reversals. Soviet physics doklady, 10(8), 707--710.

\bibitem{loshchilov2016sgdr} Loshchilov, Ilya and Hutter, Frank. (2016). SGDR: Stochastic gradient descent with warm restarts. International Conference on Learning Representations.

\bibitem{loshchilov2017decoupled} Loshchilov, Ilya and Hutter, Frank. (2017). Decoupled weight decay regularization. International Conference on Learning Representations.

\bibitem{luqmanArabsign2023} Luqman, Hamzah. (2023). ArabSign: A Multi-modality Dataset and Benchmark for Continuous Arabic Sign Language Recognition. 2023 IEEE 17th International Conference on Automatic Face and Gesture Recognition (FG).

\bibitem{mslr2025} Luqman, Hamzah and Palazzo, Simone and Alowaifeer, Maad and Mineo, Raffaele. (2025). Multimodal Sign Language Recognition (MSLR) Workshop. Proceedings of the IEEE/CVF International Conference on Computer Vision (ICCV) Workshops.

\bibitem{morris2004and} Morris, Andrew C and Maier, Viktoria and Green, Phil. (2004). From WER and RIL to MER and WIL: improved evaluation measures for connected speech recognition. Proceedings of the 8th International Conference on Spoken Language Processing.

\bibitem{park2019specaugment} Park, Daniel S and Chan, William and Zhang, Yu and Chiu, Chung-Cheng and Zoph, Barret and Cubuk, Ekin D and Le, Quoc V. (2019). SpecAugment: A Simple Data Augmentation Method for Automatic Speech Recognition. arXiv preprint arXiv:1904.08779.

\bibitem{preprocessing_keypoint_sign} {Semantic Scholar}. (2025). Preprocessing for Keypoint-Based Sign Language Recognition. Semantic Scholar.

\bibitem{pu2019iterative} Pu, Junfu and Zhou, Wengang and Li, Houqiang. (2019). Iterative alignment network for continuous sign language recognition. Proceedings of the IEEE/CVF Conference on Computer Vision and Pattern Recognition.

\bibitem{Sarhan_2023_ICCV} Sarhan, Noha and Frintrop, Simone. (2023). Unraveling a Decade: A Comprehensive Survey on Isolated Sign Language Recognition. Proceedings of the IEEE/CVF International Conference on Computer Vision (ICCV) Workshops.

\bibitem{sidig2021karsl} Sidig, Ala Addin I and Luqman, Hamzah and Mahmoud, Sabri and Mohandes, Mohamed. (2021). KArSL: Arabic Sign Language Database. ACM Transactions on Asian and Low-Resource Language Information Processing (TALLIP), 20(1), 1--19.

\bibitem{srivastava2014dropout} Srivastava, Nitish and Hinton, Geoffrey and Krizhevsky, Alex and Sutskever, Ilya and Salakhutdinov, Ruslan. (2014). Dropout: a simple way to prevent neural networks from overfitting. The journal of machine learning research, 15(1), 1929--1958.

\bibitem{thoker2021skeleton} Thoker, Fida Mohammad and Doughty, Hazel and Snoek, Cees G.M. (2021). Skeleton-Contrastive 3D Action Representation Learning. Proceedings of the 29th ACM International Conference on Multimedia (MM '21).

\bibitem{torres2014automatic} Torres-Moreno, Juan-Manuel. (2014). Automatic Text Summarization. ISTE Ltd and John Wiley \& Sons, Inc.

\bibitem{vaswani2017attention} Vaswani, Ashish and Shazeer, Noam and Parmar, Niki and Uszkoreit, Jakob and Jones, Llion and Gomez, Aidan N and Kaiser, {L}ukasz and Polosukhin, Illia. (2017). Attention is all you need. Advances in neural information processing systems.

\bibitem{who2025deafness} {World Health Organization}. (2025). Deafness and hearing loss.

\bibitem{young1994detecting} Young, S.R. (1994). Detecting misrecognitions and out-of-vocabulary words. Proceedings of ICASSP '94. IEEE International Conference on Acoustics, Speech and Signal Processing.

\end{thebibliography}
\end{document}